\documentclass[conference]{IEEEtran}
\IEEEoverridecommandlockouts
\usepackage{cite}
\usepackage{amsmath,amssymb,amsfonts}
\usepackage{algorithmic}
\usepackage{graphicx}
\usepackage{textcomp}
\usepackage{xcolor}
\usepackage{booktabs}
\usepackage[table]{xcolor}
\usepackage{makecell}
\usepackage{caption}
\usepackage{hyperref}
\usepackage{authblk}
\usepackage[dvipsnames]{xcolor}

\def\BibTeX{{\rm B\kern-.05em{\sc i\kern-.025em b}\kern-.08em
    T\kern-.1667em\lower.7ex\hbox{E}\kern-.125emX}}
\begin{document}

\title{FlowIR: Depth-driven Rectified Flow for Resolving Thermal Ambiguity in Infrared Super-Resolution}

% \author{Anonymous ICME submission}

% \author[1]{
%     Xingyue Zhu\textsuperscript{\rm 1},
%     Qinyun Mei\textsuperscript{\rm 1},
%     Zhexi Min\textsuperscript{\rm 2},
%     Yang Zou\textsuperscript{\rm 2},
%     Xingyuan Li\textsuperscript{\rm 1},
%     Zhiying Jiang\textsuperscript{\rm 3},
%     Jinyuan Liu\thanks{$^{*}$Corresponding author.}
% }

\author[1]{Xingyue Zhu}
\author[1]{Qinyun Mei}
\author[2]{Zhexi Min}
\author[2]{Yang Zou}
\author[1]{Xingyuan Li}
\author[3]{Zhiying Jiang}
\author[1]{Jinyuan Liu$^{*}$\thanks{$^{*}$Corresponding author}}

\affil[1]{Dalian University of Technology, Dalian, China}
\affil[2]{Northwestern Polytechnical University, Xi'an, China}
\affil[3]{Dalian Maritime University, Dalian, China}

\maketitle

\begin{abstract}
Distinguishing different heat sources in low-resolution infrared (IR) images is inherently difficult, as multiple thermal emitters often collapse into ambiguous and overlapping responses. Existing super-resolution methods hallucinate structures from such ambiguous 2D evidence, leading to thermal drift, boundary merging, and unreliable restoration under infrared degradations. We propose \textbf{FlowIR}, a depth-driven rectified flow framework that resolves thermal ambiguity by injecting 3D geometric cues into the generation process. Pseudo-3D depth lifting and depth-guided 3D convolutions disentangle overlapping foreground objects, while cluster-wise inter-attention suppresses cross-source feature contamination. A Thermal–Geometry Consistency Loss further enforces geometric alignment during refinement. Built upon a lightweight rectified flow backbone, FlowIR restores IR images in only a few steps, producing sharply separated, degradation-robust, and geometrically faithful high-resolution results.
\end{abstract}

\begin{IEEEkeywords}
low-level vision, infrared image super-resolution, rectified flow
\end{IEEEkeywords}

\section{Introduction}
\label{sec:intro}
Infrared (IR) imaging, which captures thermal emissions independent of visible illumination, plays a critical role in safety-critical scenarios such as surveillance, autonomous driving, and perception in adverse weather or at night. However, due to longer wavelengths and expensive sensor materials, IR cameras typically exhibit low spatial resolution and limited photon efficiency. As a result, IR images are often degraded by noise, blur, stripe artifacts, and low contrast, significantly limiting their reliability in real-world applications \cite{liu2025toward, li2024contourlet}.
% Infrared (IR) imaging, owing to its ability to capture thermal emissions independent of visible illumination, has become indispensable in safety–critical perception scenarios such as surveillance, autonomous driving, and night-time or adverse-weather vision. However, IR cameras typically operate at longer wavelengths and rely on costly sensor materials, resulting in inherently low spatial resolution and limited photon efficiency. Consequently, IR images often suffer from severe degradation, including noise, blur, stripe artifacts, and low contrast, which significantly limit their reliability and downstream utility in real-world applications.

Deep learning has substantially advanced image super-resolution (SR), from early CNN-based models with residual and dense designs\cite{kim2016accurate, tai2017image} to Transformer-based architectures exploiting long-range attention\cite{liang2021swinir,chen2023activating,zou2026toward, du2026unsupervised}, and more recently, diffusion-based methods\cite{zhang2026s3diff, wu2024seesr, zou2026hatir} leveraging strong generative priors for robust detail synthesis. Despite their success on visible images, these approaches generalize poorly to infrared data, whose low texture contrast, weak structural boundaries, and thermally dominated intensity distributions fundamentally differ from RGB imagery.
% Conventional CNN-based image super-resolution (ISR) methods demonstrated the effectiveness of learning end-to-end mappings from low- to high-resolution images, with later variants improving reconstruction fidelity through deeper residual designs and dense connectivity. Transformer-based SR models further advanced performance by exploiting attention to capture long-range dependencies. Recently, diffusion-based SR approaches have leveraged powerful generative priors to synthesize realistic details while remaining robust to complex degradations. Despite their success in the visible spectrum, these methods generalize poorly to infrared imagery, whose low texture contrast, weak structural boundaries, and thermally dominated intensity patterns fundamentally differ from RGB signals.

\begin{figure}[t]
    \centering
    \includegraphics[width=0.95\linewidth]{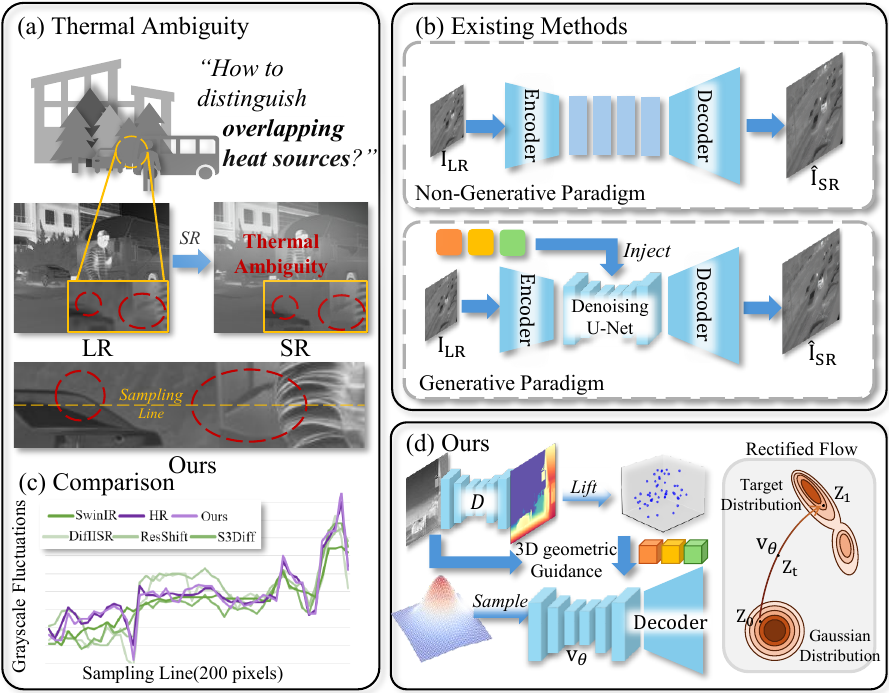}
     \caption{(a) Overlapping heat sources introduce thermal ambiguity in SR. (b) Prior methods rely solely on 2D features. (c) Grayscale fluctuations show that FlowIR better preserves pixel fidelity in overlapping heat sources regions. (d) FlowIR mitigates thermal ambiguity by integrating 3D geometric guidance into a rectified flow framework.}
    \label{fig:teaser}
    \vspace{-20pt}
\end{figure}

Recent infrared image super-resolution (IISR) methods have begun to exploit sensor–specific priors to alleviate the lack of high-frequency details. ChasNet~\cite{prajapati2021channel} boosts informative responses by channel-split convolutions, while CoRPLE~\cite{li2024contourlet} and CRG~\cite{zou2024contourlet} adopt contourlet-domain residual modeling with task-driven or prompt-based guidance. DifIISR~\cite{li2025difiisr} further integrates gradient-based alignment and perceptual priors into a diffusion framework to enhance reconstruction quality. Despite these advances, most IISR approaches are still built on purely 2D intensity cues and simplified degradations (e.g., bicubic downsampling), which limits their ability to handle complex real-world infrared degradations and severe structural ambiguity between overlapping heat sources \cite{zou2026axispose, zou2024enhancing}.

The aforementioned methods, while advancing infrared image super-resolution (IISR), still face two fundamental challenges. First, \textbf{they rely solely on 2D thermal intensities and lack geometric priors to disambiguate overlapping heat sources}, making reconstruction inherently ill-posed when multiple emitters collapse into a single low-resolution projection, as shown in Fig.~\ref{fig:teaser} (a). Second, \textbf{existing architectures and pixel-wise losses offer no mechanism to suppress cross-object feature mixing}, causing high-frequency details to leak across semantic regions—particularly under complex degradations such as blur, thermal noise, and stripe artifacts, as shown in Fig.~\ref{fig:teaser} (b). Together, these limitations lead to unstable boundaries and inconsistent structures, severely hindering reliable IISR in real-world environments.
% The aforementioned methods, while advancing IISR, still suffer from two fundamental challenges. First, they operate solely on 2D thermal intensities, lacking the geometric priors needed to disambiguate overlapping heat sources. When multiple emitters collapse into a single low-resolution projection, the reconstruction becomes inherently ill-posed, leading models to hallucinate boundaries or merge objects. Second, even when the input is not severely collapsed, existing architectures and pixel-wise losses provide no mechanism to prevent cross-object feature mixing. High-frequency cues often leak across semantic regions, especially under complex degradations such as blur, thermal noise, and stripe artifacts, leading to unstable details and inconsistent structures. These issues jointly hinder the reliability of IISR in complex environments.

To this end, we propose FlowIR, a depth-driven rectified flow framework that explicitly addresses both limitations of existing IISR methods, as shown in Fig.~\ref{fig:teaser} (d). To overcome the input-level thermal–geometric ambiguity, FlowIR introduces a Depth-Aware Point Cloud Condition that lifts depth into a pseudo-3D representation and injects explicit geometric priors into the flow trajectory, enabling reliable separation of overlapping heat sources that are indistinguishable in 2D. To mitigate model-level feature contamination, we design a 3D Cluster-Guided Decoder with intra–inter cluster attention, which enforces object-consistent representations while suppressing cross-object high-frequency leakage. In addition, a Thermal–Geometry Consistency Loss aligns the reconstructed IR image with scene geometry, ensuring physically coherent heat boundaries. These designs allow FlowIR to achieve structurally faithful and degradation-robust infrared image super-resolution in complex environments, as shown in Fig.~\ref{fig:teaser} (c). 
% To this end, we propose \textbf{FlowIR}, a depth-driven rectified flow framework that explicitly targets both limitations. To resolve the input-level thermal–geometric ambiguity, FlowIR introduces a Depth-Aware Point Cloud Condition that lifts the predicted depth map into a pseudo-3D point-cloud volume and injects it into the flow trajectory. This provides explicit geometric cues that mitigate overlapping heat sources before reconstruction, making previously ill-posed boundary recovery solvable. To handle model-level feature contamination, we design a 3D Cluster-Guided Decoder equipped with intra–inter cluster attention, which enforces feature consistency within each semantic cluster while preventing high-frequency leakage across different objects. Finally, a Thermal–Geometry Consistency Loss further constrains the SR output to remain aligned with scene geometry, ensuring physically coherent heat boundaries. Together, these components enable FlowIR to perform structurally faithful and degradation-robust IISR in complex environments. Our main contributions are manifold:

Our main contributions are manifold: 1) We propose FlowIR, a depth-driven rectified flow framework that integrates depth-derived 3D cues into rectified flow to produce structurally faithful and degradation-robust reconstructions. 2) To overcome input-level thermal–geometric ambiguity, we introduce a Depth-Aware Point Cloud Condition that lifts depth into a pseudo-3D representation and supplies explicit geometry to separate overlapping heat sources. 3) To mitigate model-level feature contamination, we design a 3D Cluster-Guided Decoder with intra–inter cluster attention and a Thermal–Geometry Consistency Loss, enforcing object-consistent features and geometrically aligned heat boundaries.

% \begin{itemize}
%     \item We propose FlowIR, a depth-driven rectified flow framework that integrates depth-derived 3D cues into rectified flow to produce structurally faithful and degradation-robust reconstructions.
%     \item To overcome input-level thermal–geometric ambiguity, we introduce a Depth-Aware Point Cloud Condition that lifts depth into a pseudo-3D representation and supplies explicit geometry to separate overlapping heat sources.
%     \item To mitigate model-level feature contamination, we design a 3D Cluster-Guided Decoder with intra–inter cluster attention and a Thermal–Geometry Consistency Loss, enforcing object-consistent features and geometrically aligned heat boundaries.
% \end{itemize}

\begin{figure*}[!tp]
    \centering
    \includegraphics[width=0.85\linewidth]{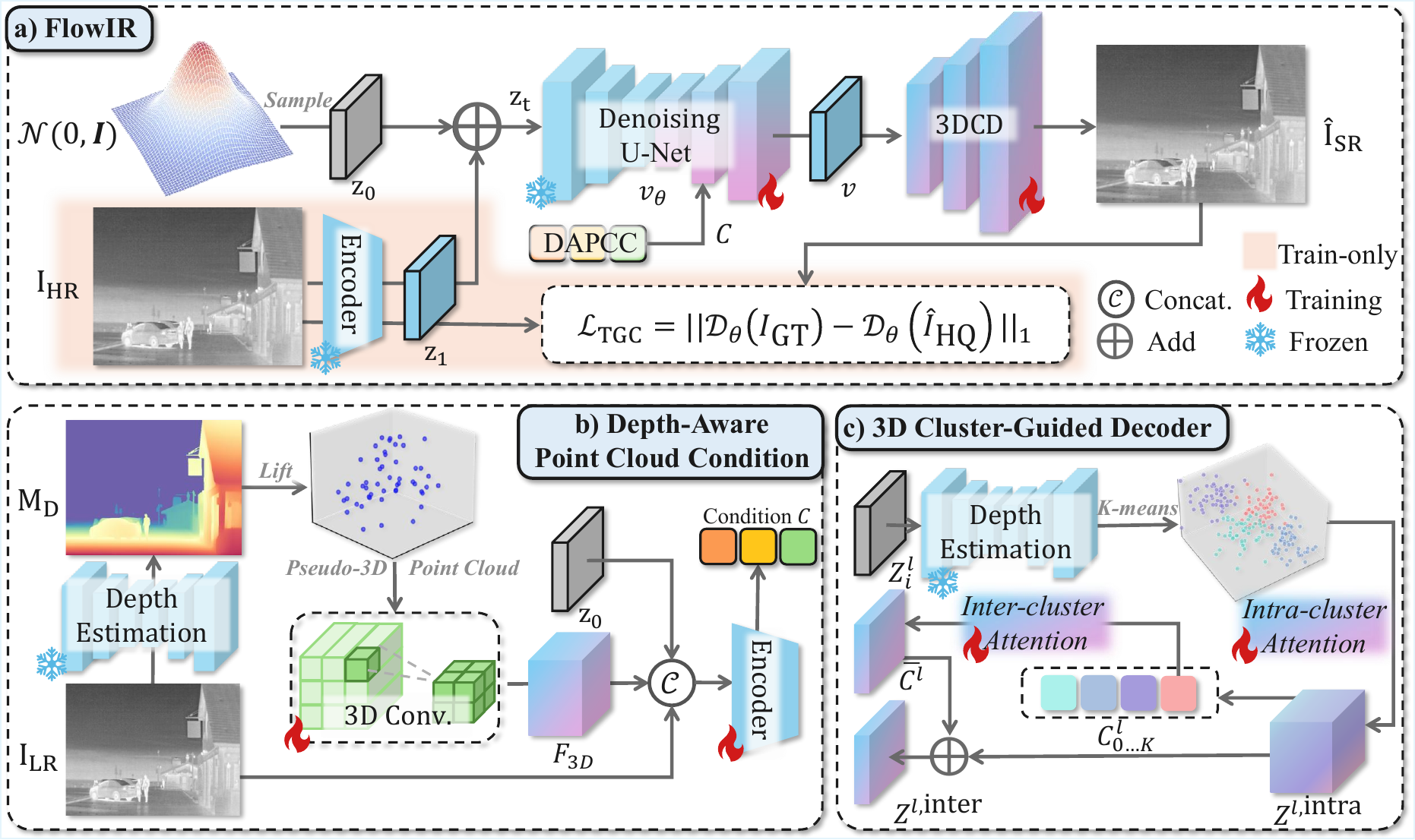}
    \caption{Overview of FlowIR. a) The overall pipeline, a depth-driven rectified flow framework that integrates depth-derived 3D cues into rectified flow to produce structurally faithful and degradation-robust reconstructions. b) The Depth-Aware Point Cloud Condition (DAPCC) module, which provides explicit geometric cues to mitigate overlapping heat sources by injecting depth-aware guidance into the flow trajectory. c) The 3D Cluster-Guided Decoder (3DCD) that enforces feature consistency within each semantic cluster while preventing high-frequency leakage across different objects.}
    \vspace{-10pt}
    \label{fig:pipeline}
\end{figure*}

\section{Method}
\subsection{Overview}
As shown in Fig.~\ref{fig:pipeline}, the process begins by sampling a noisy image from a standard Gaussian distribution. In the Depth-Aware Point Cloud Condition (DAPCC) module, the degraded infrared image is utilized to generate a condition enriched with 3D geometric information at the input level. This condition then guides the estimation of the rectified flow. Subsequently, the noisy image is fed into a pre-trained U-Net, which iteratively samples under DAPCC guidance to estimate the rectified flow and derive the restored infrared image features. Finally, the 3D Cluster-Guided Decoder (3DCD) is employed to reconstruct a restored infrared image with thermal source consistency.

\subsection{Depth-Aware Point Cloud Condition}
To tackle the thermal–geometric ambiguity in degraded infrared images, we propose the Depth-Aware Point Cloud Condition, which utilizes point clouds to infuse 3D geometric constraints into the estimation of the rectified flow trajectory. While previous image restoration methods based on rectified flow also utilize the original low-quality image as a condition, they rely solely on its 2D image features. Degraded infrared images often lack high-frequency details, and infrared images inherently suffer from blurred thermal boundaries. This can lead to incorrect restoration of thermal edges, further compromising downstream tasks such as detection and segmentation. Our Depth-Aware Point Cloud Condition addresses this by explicitly utilizing 3D geometric information to guide the reconstruction pathway, thereby enhancing the restoration reliability and improving the thermal consistency of the results.

During training, the condition $C$ for time step $t$ is constructed as follows. Initially, the low-quality infrared image $\text{I}_\text{LR} \in \mathbb{R}^{H \times W}$ is passed through a pre-trained depth estimation network $\mathcal{D}$, and the predicted depth map $M_{D}$ is projected into a pseudo-3D point-cloud volume $V_{pc} \in \mathbb{R}^{H \times W \times Z}$. Subsequently, a 3D convolutional module is employed to extract 3D geometric features $F_{3D}\in \mathbb{R}^{H \times W \times C'}$ from the infrared scene. These features, together with $\text{I}_\text{LR}$ and the noisy image $z_t$ at step $t$, are then concatenated and passed through a fully connected layer to generate the final condition. This process can be expressed as:

\begin{equation}
    M_D = \mathcal{D}_\theta(\text{I}_\text{LR}), \quad V_{pc} = \mathcal{P}(M_d, \text{I}_\text{LR}),
\end{equation}
\begin{equation}
    F_{3D} = \text{Conv3D}(V_{pc}),
\end{equation}
\begin{equation}
    C_t = \operatorname{MLP}\big(\operatorname{Concat}(\text{I}_\text{LR},\; F_{3D},\; z_t)\big),
\end{equation}

\noindent where $\mathcal{P}$ represents the lifting process. The rectified flow $v_\theta$ is then estimated with $C$ being injected into the decoder of the U-Net. Subsequently, the restored feature $z_1$ can be computed and then serve as the input to the 3DCD.

\subsection{3D Cluster-Guided Decoder}
Infrared images often lack texture and structural information due to physical factors such as heat diffusion, thermal noise, and stripe artifacts. When further degraded by compression-induced detail loss, direct reconstruction tends to produce unstable details and structurally inconsistent results, leading to the accumulation of errors at the feature level during decoding.

To address this issue, we propose the 3D Cluster-Guided Decoder. During feature reconstruction, it leverages depth information to maintain thermal consistency through intra-cluster self-attention, while promoting global information flow via inter-cluster cross-attention. 

In real-world infrared scenes, objects of the same semantic category typically exhibit continuous depth and similar texture characteristics. Building on this observation, we propose the Intra-Inter Cluster Attention (I\(^{2}\)CA) mechanism. By incorporating K-Means clustering, I\(^{2}\)CA establishes a hierarchical attention framework that ensures feature consistency within clusters and contextual correlation across clusters, thereby mitigating thermal boundary instability during infrared image restoration.

Given the input $Z^l$ to the l-th upsample layer of the decoder, the features are first transformed into a pseudo-3D point cloud volume using the depth map obtained from a depth estimation network. Then, K-means clustering is applied to group points that share similar spatial and semantic characteristics into $K$ clusters. 
For each cluster $k$, an inter-cluster mask $M_k$ is computed as $M_k^l = \{ i \in [1,N] \mid L_{i}^l = k\}$. This mask is applied to $Z^l$ to extract the corresponding cluster-specific features $Z_{k}^l = \{ Z_{i}^l \mid i \in M_{k}^l \} \in \mathbb{R}^{M_{k}^l \times C}$. Subsequently, intra-cluster self-attention is performed on these features to model internal dependencies, yielding the refined features after intra-cluster interaction $Z^{l,\text{intra}}_k=\text{SelfAttn}(Z_{k}^l)$.

After obtaining the features following inter-cluster interaction, we compute a comprehensive representation $C^l = [C_{0}^l, C_{1}^l, \dots, C_{K-1}^l] \in \mathbb{R}^{K \times C}$ for each class via averaging, and then apply global inter-class attention to these aggregated representations.
\begin{equation}
    C_{k}^l = \frac{1}{M_{k}^l} \sum_{i \in M_{k}^l} Z_{i}^{l, \text{intra}},
\end{equation}
\begin{equation}
    \begin{aligned}
    Q^l = C^l W^Q, \quad K^l = C^l& W^K, \quad V^l = C^l W^V \\
    \bar{C^{l}} = \operatorname{softmax}&\left( \frac{Q^l K^{l\top}}{\sqrt{d_h}} \right) V^l,
    \end{aligned}
\end{equation}

\begin{equation}
    Z^{l,\text{inter}} = Z^{l,\text{intra}} + \bar{C^{l}}.
\end{equation}

Finally, for each Upsample block in the decoder, the features are updated as follows before upsampling:
\begin{equation}
    Z^{l+1} = \operatorname{LayerNorm}(Z^{l,\text{inter}} + \operatorname{FFN}(Z^{l,\text{inter}})),
\end{equation}

By incorporating 3D geometric priors at the feature level, the 3D Cluster-Guided Decoder facilitates interaction between depth-aware geometric information and infrared texture features, ultimately yielding a high-fidelity infrared image $ \hat{\text{I}}_\text{SR} $ that maintains fidelity in both thermal semantics and geometric structure.

\begin{table*}[!ht]
\centering
\caption{Quantitative comparison on FMB and \(\mathrm{M^{3}FD}\) (\(\times\)4). The best is in \textbf{bold}, while the second is \underline{underlined}. For FMB, Set5/10/20 are randomly sampled subsets. For \(\mathrm{M^{3}FD}\), the three sets correspond to three degradation levels of increasing severity.}
\small
\resizebox{\textwidth}{!}{
\begin{tabular}{@{\hspace{5pt}}lccc@{\hspace{4pt}}c@{\hspace{4pt}}cccc@{\hspace{4pt}}c@{\hspace{4pt}}cccc@{\hspace{4pt}}c@{\hspace{4pt}}c@{\hspace{5pt}}}
\Xhline{1.1pt}
\multicolumn{1}{c}{\textbf{Datasets}} 
& \multicolumn{5}{c}{\textbf{Set5}} 
& \multicolumn{5}{c}{\textbf{Set10}} 
& \multicolumn{5}{c}{\textbf{Set20}} \\
\cmidrule(lr){1-1} \cmidrule(lr){2-6} \cmidrule(lr){7-11} \cmidrule(lr){12-16}
 
\multicolumn{1}{c}{\textbf{Methods}}
& PSNR$\uparrow$ & SSIM$\uparrow$ & LPIPS$\downarrow$ &  CLIP-IQA$\uparrow$ & MUSIQ$\uparrow$
& PSNR$\uparrow$ & SSIM$\uparrow$ & LPIPS$\downarrow$ &  CLIP-IQA$\uparrow$ & MUSIQ$\uparrow$ 
& PSNR$\uparrow$ & SSIM$\uparrow$ & LPIPS$\downarrow$ &  CLIP-IQA$\uparrow$ & MUSIQ$\uparrow$ \\
\hline

\multicolumn{16}{c}{\textbf{FMB}} \\
\hline
ChasNet   & 42.2995 & 0.9732 & 0.1587 & 0.1432 & 27.9576 & 42.0907 & 0.9429 & 0.1607 & 0.1436 & 28.3499 & 42.0137 & 0.9713 & 0.1612 & 0.1445 & 28.1572 \\
DAT       & \underline{42.4312} & 0.9740 & 0.1587 & 0.1462 & 30.3660 & 42.3480 & \underline{0.9713} & 0.1566 & 0.1435 & 30.0319 & \underline{42.4958} & \textbf{0.9777} & 0.1560 & 0.1508 & 30.3948 \\
DifIISR   & 33.3211 & 0.9073 & 0.1780 & 0.3414 & 46.2157 & 33.0764 & 0.9462 & 0.1812 & 0.3463 & 46.4277 & 34.4098 & 0.9461 & 0.1761 & 0.3381 & 45.9119  \\
HAT       & 42.4071 & \underline{0.9741} & 0.1590 & 0.1490 & 29.6160 & \underline{42.7815} & 0.9677 & 0.1569 & 0.1489 & 29.9870 & 42.0661 & 0.9428 & 0.1582 & 0.1527 & 29.9735 \\
OSEDiff   & 29.8344 & 0.9142 & 0.1493 & 0.1255 & 26.7615 & 29.9214 & 0.9028 & 0.1462 & 0.1233 & 27.0928 & 28.8406 & 0.9403 & 0.1521 & 0.1229 & 26.8410 \\
ResShift  & 30.9180 & 0.8606 & 0.2345 & 0.2569 & 41.8227 & 30.5916 & 0.8617 & 0.2348 & 0.2583 & 41.4749 & 30.9579 & 0.8843 & 0.2340 & 0.2579 & 41.8641 \\
S3Diff    & 35.9839 & 0.9367 & \textbf{0.0987} & 0.1273 & 27.6296 & 36.1984 & 0.9003 & \underline{0.0950} & 0.1238 & 27.4848 & 37.0308 & 0.9447 & \underline{0.0973} & 0.1306 & 27.7889 \\
SeeSR     & 32.1978 & 0.8918 & 0.2080 & \underline{0.3662} & \underline{54.0856} & 31.7994 & 0.8570 & 0.2112 & \underline{0.3625} & \underline{54.3989} & 32.5235 & 0.8907 & 0.2085 & \underline{0.3649} & \underline{54.1967} \\
SinSR     & 33.3448 & 0.8996 & 0.1856 & 0.3598 & 46.7329 & 33.3051 & 0.9315 & 0.1873 & 0.3622 & 46.9874 & 34.4848 & 0.9044 & 0.1891 & 0.3612 & 46.8769 \\
SwinIR    & 33.9122 & 0.9340 & 0.1701 & 0.0800 & 27.8900 & 34.0223 & 0.9168 & 0.1662 & 0.0753 & 27.9453 & 33.7240 & 0.9499 & 0.1701 & 0.0827 & 28.2644 \\
TSD-SR    & 33.4534 & 0.9090 & 0.1409 & 0.1036 & 29.0244 & 33.0868 & 0.9460 & 0.1426 & 0.1037 & 29.4242 & 32.1092 & 0.9409 & 0.1391 & 0.1017 & 28.6661 \\
\rowcolor[gray]{0.9}
Ours      & \textbf{45.9893} & \textbf{0.9743} & \underline{0.0993} & \textbf{0.3711} & \textbf{56.1121} & \textbf{45.8921} & \textbf{0.9783} & \textbf{0.0939} & \textbf{0.3820} & \textbf{55.5468} & \textbf{46.1166} & \underline{0.9767} & \textbf{0.0961} & \textbf{0.3747} & \textbf{54.3354} \\
\hline
\multicolumn{16}{c}{\textbf{\boldmath$\mathrm{M^{3}FD}$}} \\ 
\hline
ChasNet  & 26.4945 & 0.6635 & 0.5922 & 0.2740 & 22.1305 
                & 26.2028 & 0.6531 & 0.6067 & 0.2607 & 21.0788 
                & 25.4257 & 0.6366 & 0.6156 & 0.2410 & 20.9368  \\
DAT       & 26.9818 & 0.6703 & 0.5930 & 0.3114 & 24.2585 
               & 26.1590 & 0.6457 & 0.6088 & 0.3070 & 23.5410 
               & 25.2609 & 0.6285 & 0.6296 & 0.2877 & 21.6303 \\
DifIISR   & 27.6127 & 0.7071 & 0.4540 & \underline{0.5486} & 38.8221 
                & 27.3071 & 0.6953 & 0.4635 & \underline{0.5473} & 37.8671 
                & 26.7638 & 0.6659 & 0.4771 & \underline{0.5384} & 37.0030  \\
HAT       & 27.0783 & 0.6781 & 0.5903 & 0.3514 & 21.7532 
                & 26.7508 & 0.6758 & 0.6043 & 0.3337 & 21.6000 
                & 25.8480 & 0.6696 & 0.6129 & 0.3294 & 20.6502 \\
OSEDiff   & 27.0964 & 0.8416 & 0.2699 & 0.1746 & 30.2667 
                & 26.4260 & \textbf{0.8392} & 0.2712 & 0.1603 & 29.3309 
                & 26.1743 & \textbf{0.8277} & 0.2906 & 0.1599 & 28.8840 \\
ResShift  & 26.7873 & 0.6454 & 0.5536 & 0.3494 & 30.8375 
                & 25.7273 & 0.6439 & 0.5618 & 0.3477 & 29.1958 
                & 25.0640 & 0.6377 & 0.5753 & 0.3208 & 28.4651 \\
S3Diff    & \underline{29.0797} & \underline{0.8433} & \underline{0.2446} & 0.2035 & 32.6686 
                & \textbf{28.5746} & 0.8310 & \underline{0.2501} & 0.2089 & 31.8038
                & \underline{28.2000} & 0.8177 & \underline{0.2589} & 0.1900 & 31.0597  \\
SeeSR     & 27.3091 & 0.7871 & 0.3529 & 0.4860 & \underline{53.9075}
                & 27.0611 & 0.7713 & 0.3683 & 0.4752 & \underline{51.8196} 
                & 26.5842 & 0.7634 & 0.3813 & 0.4610 & \underline{50.7902} \\
SinSR     & 27.2619 & 0.6779 & 0.4785 & 0.5313 & 38.3171
                & 26.3434 & 0.6703 & 0.4870 & 0.5196 & 37.4036
                & 25.6494 & 0.6638 & 0.5043 & 0.4984 & 36.6088 \\
SwinIR    & 26.9230 & 0.6769 & 0.5814 & 0.3236 & 27.7717 
                & 26.7383 & 0.6676 & 0.5923 & 0.3198 & 28.1318
                & 26.2332 & 0.6627 & 0.5980 & 0.2992 & 25.5470 \\
TSD-SR    & 28.4807 & 0.8285 & 0.2809 & 0.1505 & 28.9007 
                & 27.8437 & 0.8237 & 0.2913 & 0.1360 & 27.8249 
                & 26.4642 & 0.8150 & 0.3045 & 0.1162 & 26.8972 \\
\rowcolor[gray]{0.9}
Ours      & \textbf{30.5532} & \textbf{0.8916} &\textbf{ 0.2060} & \textbf{0.5599} & \textbf{55.6257} 
                & \underline{28.3384} & \underline{0.8274} & \textbf{0.2227} & \textbf{0.5498} & \textbf{53.0167} 
                & \textbf{28.4713} & \underline{0.8216 }& \textbf{0.2379} & \textbf{0.5431} &\textbf{ 52.9505} \\
\Xhline{1.1pt}
\end{tabular}
}
\label{tab:final_metrics_layout}
\vspace{-10pt}
\end{table*}

\subsection{Optimization}
During model training, we aim to leverage the rich priors of diffusion models while adapting them for infrared image super-resolution. To this end, we freeze the U‑Net and Encoder, and train only the Depth-Aware Point Cloud Condition generator, the injection modules in the U‑Net, and the 3D Cluster-Guided Decoder. While conventional super‑resolution losses are constrained solely in the 2D image space (e.g., focusing on improving overall visual quality and pixel‑level reconstruction), they lack explicit consideration of the 3D scene structure. To ensure that the super‑resolution process preserves the 3D boundaries and spatial relationships of thermal objects, we introduce the Thermal-Geometry Consistency Loss as 
\begin{equation}
    \mathcal{L}_{\text{TGC}} =  \left\|\mathcal{D}_\theta(\text{I}_\text{GT})-\mathcal{D}_\theta(\hat{\text{I}}_{\text{SR}})\right\|_1.
\end{equation}

The Thermal-Geometry Consistency Loss enforces consistency between the restored HR image and the ground truth in the depth estimation space, thereby constraining the network to learn physically plausible super‑resolution reconstructions. 

To further enhance the supervision of thermal source segmentation and achieve sharper boundaries, we introduce a Thermal Boundary Loss as $\mathcal{L}_{\text{TB}} = \left\| \nabla \mathcal{D}_\theta(\text{I}_\text{GT})-\nabla\mathcal{D}_\theta(\hat{\text{I}}_\text{SR})\right\|_1$, where $\nabla(\cdot)$ denotes the Laplacian operator. The overall training loss is formulated by combining the Thermal-Geometry Consistency Loss and the Thermal Boundary Loss with a pixel-wise L1 loss between $\text{I}_{\text{GT}}$ and $\hat{\text{I}}_\text{SR}$.

\section{Experiments}
\subsection{Experimental Settings}
\noindent \textbf{Implementation Details.} 
We initialize FlowIR with a pre-trained Stable Diffusion model to leverage its strong generative prior. During training, the U-Net and encoder are frozen, while only the Depth-Aware Point Cloud Condition generator, the depth injection modules, and the 3D Cluster-Guided Decoder are optimized. For depth estimation, we used pre-trained Depth Anything 3 \cite{depthanything3}. All experiments are conducted on an NVIDIA A100 GPU using the Adam optimizer, with 8 clusters and 5 sampling steps.
% To leverage the powerful prior of diffusion models, we first initialize the path predictor using a pre-trained Stable Diffusion model. We then freeze the U-Net and encoder, and train only the Depth-Aware Point Cloud Condition generator, the injection modules within the U-Net, and the 3D Cluster-Guided Decoder. Training is conducted on an NVIDIA A100 GPU using the Adam optimizer, with 8 clusters and 5 sampling steps.

\noindent \textbf{Datasets and Evaluation Metrics.} 
To improve robustness, we adopt the real-world infrared degradation strategy from ControlFusion \cite{tang2025controlfusion}, which synthesizes both single and hybrid degradations. Degradations are applied to the FMB \cite{liu2023multi} dataset with random intensities and the \(\mathrm{M^{3}FD}\) \cite{liu2022target} datasets at three severity levels, and both datasets then underwent \(\times\)4 downsampling. The degraded FMB dataset is split into 70\% training and 30\% testing, while evaluation is performed on both the FMB test set and the full degraded \(\mathrm{M^{3}FD}\) dataset. Performance is assessed using five standard metrics: PSNR, SSIM, LPIPS, CLIP-IQA, and MUSIQ, covering both reconstruction fidelity and perceptual quality.
% To enhance the model's robustness, we utilize the Infrared Degradation Processing technique from HM-TIR, which is designed to generate both single and hybrid degradation. Specifically, we introduce degradations with randomly varying intensities to the FMB dataset, and apply three distinct levels of degradation (from weak to strong) to the $\mathrm{M^{3}FD}$ dataset. For training, we randomly split the degraded FMB dataset into 70\% for training and 30\% for testing. During the evaluation phase, we test our model on both this degraded FMB test set and the entire degraded $\mathrm{M^{3}FD}$ dataset. To comprehensively assess both fidelity and perceptual quality, we report five widely used metrics: Peak Signal-to-Noise Ratio (PSNR), Structural Similarity Index Measure (SSIM), Learned Perceptual Image Patch Similarity (LPIPS), CLIP for Image Quality Assessment (CLIP), and Multi-scale Image Quality Transformer (MUSIQ).

\begin{figure*}[!h]
    \centering
    \includegraphics[width=0.95\textwidth]{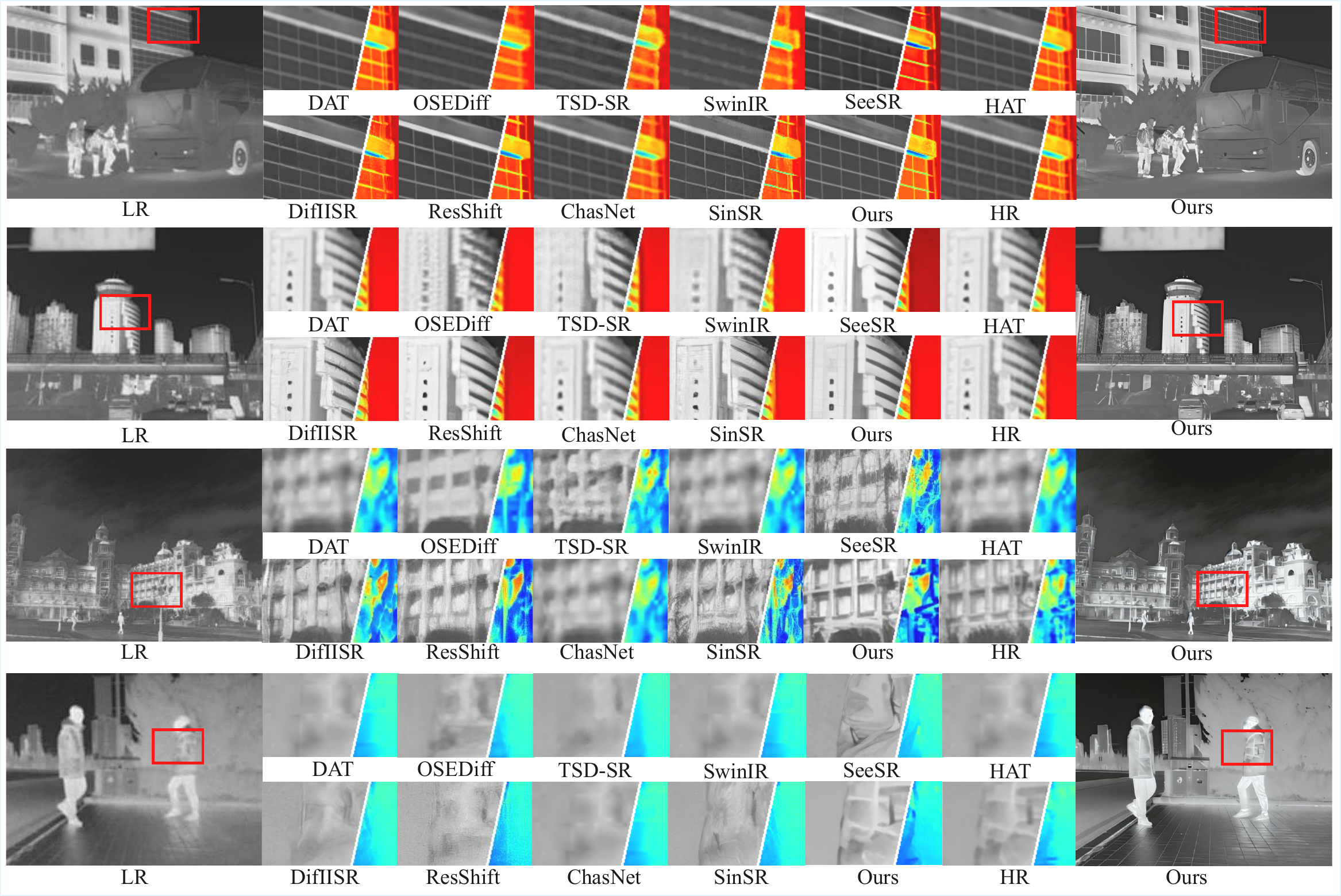}
    \caption{Qualitative results. The first and second rows are from the static scenes of the FMB dataset (\(\times\)4), while the third and fourth rows are from the $\mathrm{M^3FD}$ dataset (\(\times\)4).}
    \label{fig:Comparision}
    \vspace{-17pt}
\end{figure*}

\noindent \textbf{Comparative Methods.} We conduct a comprehensive comparison with 11 recent state-of-the-art image super-resolution methods, including those specifically designed for infrared images. The comparison methods encompass both non-generative paradigms like ChasNet\cite{prajapati2021channel}, DAT\cite{chen2023dual}, HAT\cite{chen2023activating}, and generative paradigms like DifIISR\cite{li2025difiisr}, OSEDiff\cite{wu2024one}, ResShift\cite{yue2023resshift}, S3Diff\cite{zhang2026s3diff}, SeeSR\cite{wu2024seesr}, SinSR\cite{wang2024sinsr}, and TSD-SR\cite{dong2025tsd}.

\begin{figure}[t]
    \centering
    \includegraphics[width=0.85\linewidth]{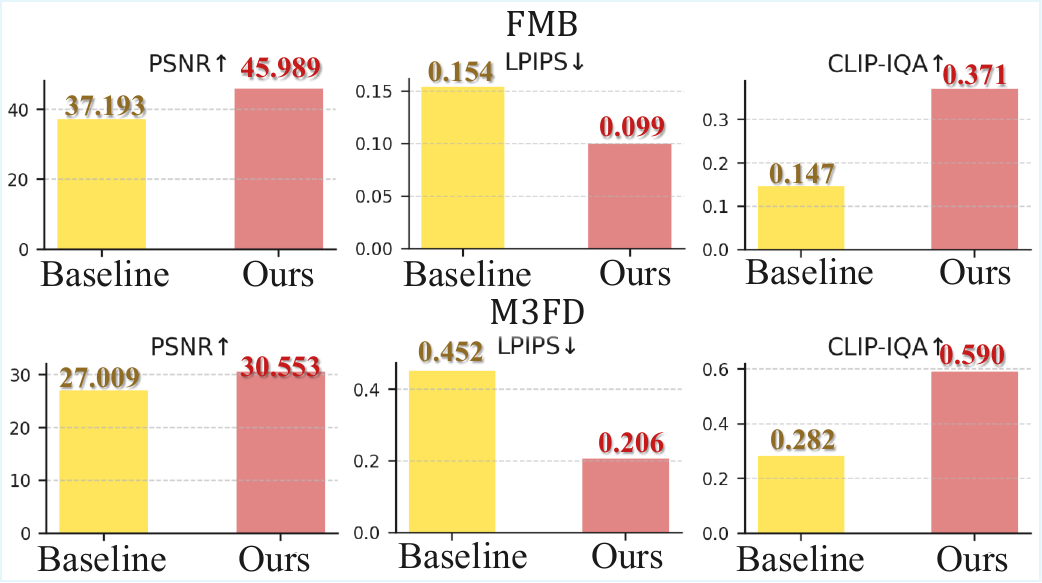}
     \caption{Quantitative ablation study on Depth Awareness.}
    \label{fig:abl_depth_quantity}
    \vspace{-15pt}
\end{figure}
\begin{figure}[t]
    \centering
    \includegraphics[width=0.85\linewidth]{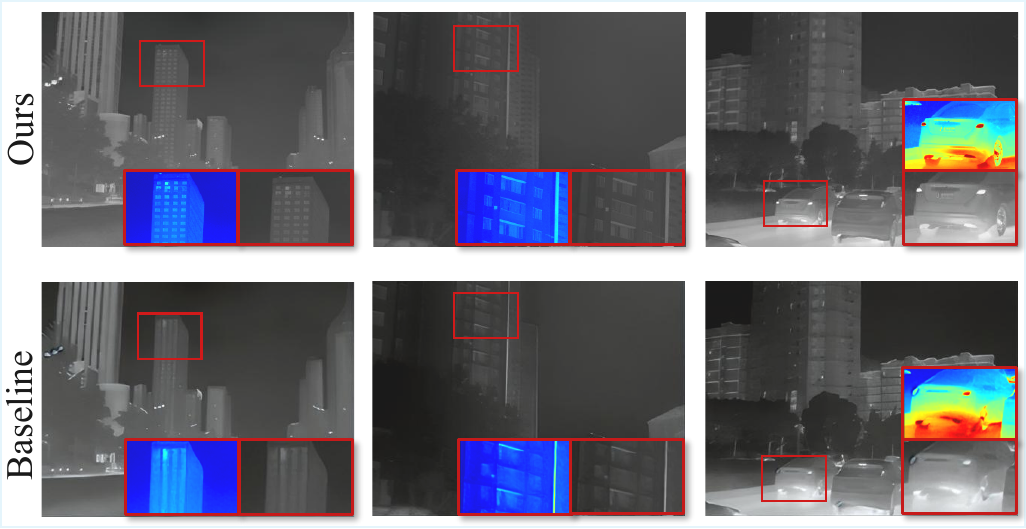}
     \caption{Qualitative ablation study on Depth Awareness.}
    \label{fig:abl_depth_quality}
    \vspace{-18pt}
\end{figure}

\subsection{Quantitative Comparison}
Table~\ref{tab:final_metrics_layout} reports quantitative super-resolution results on the FMB and $\mathrm{M^{3}FD}$ datasets. Results on FMB are averaged over multiple randomly sampled subsets, while $\mathrm{M^{3}FD}$ is evaluated under three degradation levels of increasing severity. Our method consistently outperforms all generative baselines in PSNR and SSIM on both datasets. Moreover, it achieves the best performance on perception-oriented metrics across nearly all test settings, except for FMB. These results indicate that FlowIR delivers high-quality reconstructions in both human perception and deep feature space, validating the effectiveness of depth integration at multiple stages of the framework.
% Table~\ref{tab:final_metrics_layout} presents a quantitative comparison of super-resolution performance on the FMB and $\mathrm{M^{3}FD}$ datasets. For FMB, we report metrics averaged over multiple randomly selected subsets. For $\mathrm{M^{3}FD}$, we evaluate performance across three degradation levels of increasing intensity.

% The results demonstrate that our method outperforms all generative counterparts in terms of PSNR and SSIM on both datasets, indicating superior pixel-wise fidelity. Notably, our approach also achieves the best scores on the CLIP-IQA, LPIPS, and MUSIQ metrics across almost all test sets, with the sole exception of one subset in FMB. This consistent advantage across perception-oriented metrics confirms that our results maintain high quality not only in human perception but also at the deep feature level, validating the effectiveness of integrating depth information at three distinct levels of our framework.

\subsection{Qualitative Results}
Figure~\ref{fig:Comparision} visualizes super-resolution results of four infrared images from the FMB (top) and $\mathrm{M^{3}FD}$ (bottom) datasets. As shown, non-generative methods struggle to recover sharp and reliable thermal boundaries, as they rely solely on 2D infrared features. Their performance further degrades under stronger degradations, often failing to separate foreground and background objects with similar temperatures. Generative approaches produce perceptually sharper results, while the lack of explicit 3D geometric guidance leads to distorted boundaries and noticeable artifacts. In contrast, our method consistently reconstructs sharp, well-separated thermal structures that closely align with the ground truth.
% To visually demonstrate the effectiveness of our method for IISR, Fig.~\ref{fig:Comparision} presents the super-resolution results of four infrared images processed by different methods. The top two samples are from the FMB dataset, while the bottom two are from the $\mathrm{M^{3}FD}$ dataset.

% As shown in the figures, the non-generative methods (i.e., SinSR, DAT, HAT, and ChasNet) fail to recover high-quality, sharp thermal boundaries. This limitation stems from their reliance only on 2D planar features from the infrared images. Their performance degrades further with increased degradation intensity, often failing to distinguish between foreground and background objects of similar temperature. The generative methods—DifIISR, OSEDiff, ResShift, S3Diff, SeeSR, SinSR, and TSD-SR—produce results with perceptually sharper edges; the absence of direct 3D geometric guidance leads to distorted thermal boundaries and noticeable artifacts in their outputs. In contrast, our method not only reconstructs sharp thermal boundaries but also produces structures that are highly consistent with ground truth.

\subsection{Ablation Studies}

\subsubsection{Depth Awareness}
To evaluate the impact of depth information in IISR, we compare FlowIR with its baseline \cite{zhu2024flowie} variant. As shown in Fig.~\ref{fig:abl_depth_quantity}, incorporating depth cues yields consistent performance gains on both the FMB and $\mathrm{M^{3}FD}$ datasets. In particular, the CLIP-IQA score improves significantly, indicating a substantial enhancement in perceptual quality. The qualitative comparisons in Fig.~\ref{fig:abl_depth_quality} further corroborate these results.
% To validate the effectiveness of depth information in IISR, we compare our proposed method against our baseline model. As shown in Fig.~\ref{fig:abl_depth_quantity}, our full model achieves consistent performance gains on both the FMB and M3FD datasets through this comprehensive, multi-level utilization of depth cues. Notably, the CLIP score shows a significant improvement of 0.224 and 0.308 on the respective datasets, highlighting a substantial enhancement in perceptual image quality. We provide three sets of visual comparisons in Fig.~\ref{fig:abl_depth_quality}, which align with the quantitative improvements. Together, the three sample groups highlight the capability of depth awareness in recovering detailed image structures and addressing thermal ambiguity.

\subsubsection{Individual Modules}
\begin{figure}[t]
    \centering
    \includegraphics[width=1\linewidth]{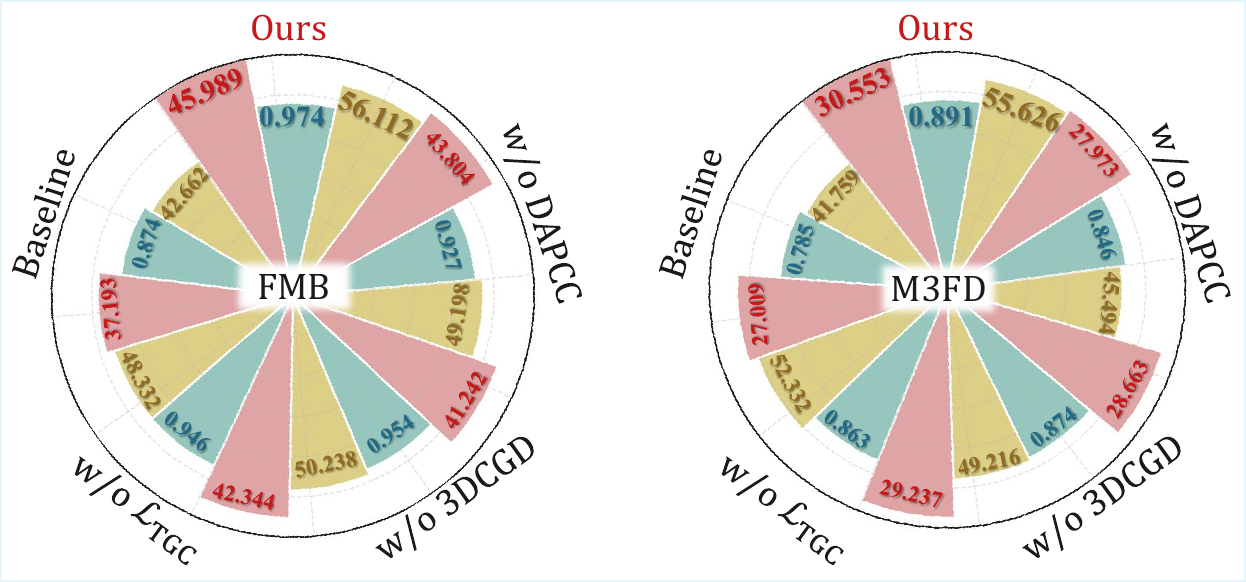}
     \caption{Quantitative ablation study of individual modules on \textcolor{red}{PSNR}, \textcolor{teal}{SSIM}, \textcolor{Goldenrod}{MUSIQ}.}
    \label{fig:abl_module_quantity}
    \vspace{-15pt}
\end{figure}

\begin{figure}[t]
    \centering
    \includegraphics[width=0.9\linewidth]{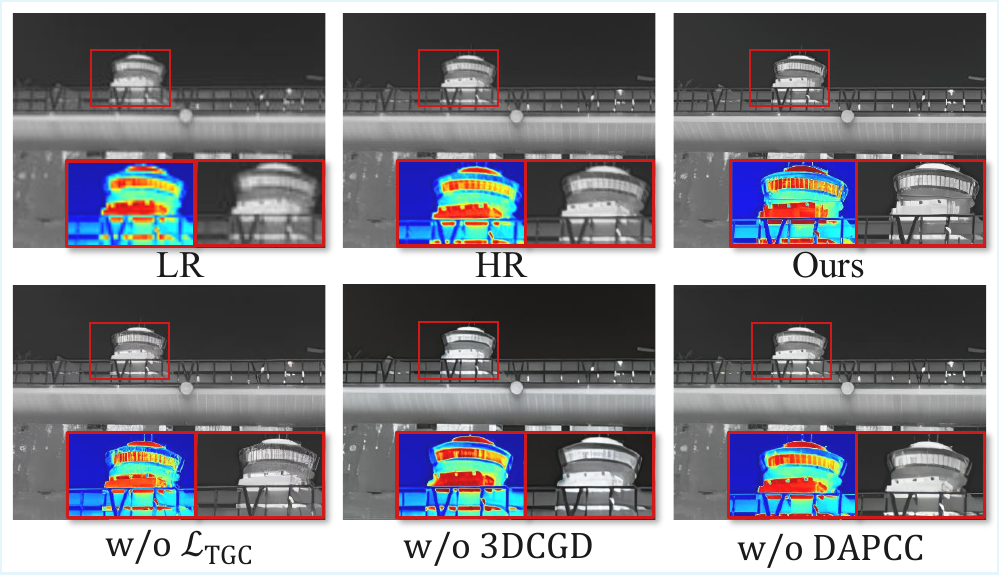}
     \caption{Qualitative ablation study on individual modules.}
    \label{fig:abl_module_quality}
    \vspace{-18pt}
\end{figure}

To evaluate the contribution of each component, we perform ablation studies by individually removing the DAPCC, the 3DCGD, and the ($\mathcal{L}_{\text{TGC}}$), followed by separate training and evaluation. As shown in Fig.~\ref{fig:abl_module_quantity}, removing DAPCC causes a pronounced drop in PSNR on both datasets, indicating that depth-guided flow prediction is critical for pixel-wise fidelity. In contrast, removing either 3DCGD or $\mathcal{L}_{\text{TGC}}$ mainly degrades SSIM and MUSIQ. The qualitative results in Fig.~\ref{fig:abl_module_quality} further confirm that the complete FlowIR model best suppresses thermal ambiguity and structural hallucination.
% To assess the independent contribution of each depth-aware module, we conduct ablation studies by individually removing the Depth-Aware Point Cloud Condition generator (DAPCC), the 3D Cluster-Guided Decoder (3DCGD), and the Thermal–Geometry Consistency Loss ($\mathcal{L}_{\text{TGC}}$), followed by separate training and testing. As illustrated in Fig.~\ref{fig:abl_module_quantity}, removing DAPCC leads to a significant drop in PSNR on both datasets. This indicates that guiding the diffusion path prediction with depth information is crucial for achieving pixel-wise fidelity. The removal of either 3DCGD or $\mathcal{L}_{\text{TGC}}$ results in a notable decline in SSIM and MUSIQ scores. This underscores that incorporating depth information at the feature reconstruction and supervision levels is essential to recover accurate structures and high visual quality. Visual comparisons in Fig.~\ref{fig:abl_module_quality} further confirm these findings. It can be observed that the complete FlowIR model delivers the best performance in suppressing thermal ambiguity and structural hallucinations, thereby validating the collective effectiveness of the DAPCC, 3DCGD, and $\mathcal{L}_{\text{TGC}}$ components.

\section{Conclusion}
To address the thermal ambiguity in infrared image super-resolution caused by the collapse of overlapping heat sources, we propose FlowIR, a depth-driven rectified flow framework that injects explicit 3D geometric cues into the generation process. By jointly leveraging pseudo-3D depth guidance and cluster-aware feature reconstruction, FlowIR effectively resolves overlapping heat sources and stabilizes thermal boundaries. Extensive experiments validate its superiority in reconstruction fidelity under complex degradations.

\bibliographystyle{IEEEbib}
\bibliography{icme2026references}

@inproceedings{prajapati2021channel,
  title={Channel split convolutional neural network (ChaSNet) for thermal image super-resolution},
  author={Prajapati, Kalpesh and Chudasama, Vishal and Patel, Heena and Sarvaiya, Anjali and Upla, Kishor P and Raja, Kiran and Ramachandra, Raghavendra and Busch, Christoph},
  booktitle={Proceedings of the IEEE/CVF Conference on Computer Vision and Pattern Recognition},
  pages={4368--4377},
  year={2021}
}

@inproceedings{li2024contourlet,
  title={Contourlet residual for prompt learning enhanced infrared image super-resolution},
  author={Li, Xingyuan and Liu, Jinyuan and Chen, Zhixin and Zou, Yang and Ma, Long and Fan, Xin and Liu, Risheng},
  booktitle={European Conference on Computer Vision},
  pages={270--288},
  year={2024},
  organization={Springer}
}

@inproceedings{li2025difiisr,
  title={Difiisr: A diffusion model with gradient guidance for infrared image super-resolution},
  author={Li, Xingyuan and Wang, Zirui and Zou, Yang and Chen, Zhixin and Ma, Jun and Jiang, Zhiying and Ma, Long and Liu, Jinyuan},
  booktitle={Proceedings of the Computer Vision and Pattern Recognition Conference},
  pages={7534--7544},
  year={2025}
}

@inproceedings{chen2023dual,
  title={Dual aggregation transformer for image super-resolution},
  author={Chen, Zheng and Zhang, Yulun and Gu, Jinjin and Kong, Linghe and Yang, Xiaokang and Yu, Fisher},
  booktitle={Proceedings of the IEEE/CVF international conference on computer vision},
  pages={12312--12321},
  year={2023}
}

@inproceedings{chen2023activating,
  title={Activating more pixels in image super-resolution transformer},
  author={Chen, Xiangyu and Wang, Xintao and Zhou, Jiantao and Qiao, Yu and Dong, Chao},
  booktitle={Proceedings of the IEEE/CVF conference on computer vision and pattern recognition},
  pages={22367--22377},
  year={2023}
}

@article{wu2024one,
  title={One-step effective diffusion network for real-world image super-resolution},
  author={Wu, Rongyuan and Sun, Lingchen and Ma, Zhiyuan and Zhang, Lei},
  journal={Advances in Neural Information Processing Systems},
  volume={37},
  pages={92529--92553},
  year={2024}
}

@article{yue2023resshift,
  title={Resshift: Efficient diffusion model for image super-resolution by residual shifting},
  author={Yue, Zongsheng and Wang, Jianyi and Loy, Chen Change},
  journal={Advances in Neural Information Processing Systems},
  volume={36},
  pages={13294--13307},
  year={2023}
}

@article{zhang2026s3diff,
  title={S3diff: Semantic Fusion and Structure-Guided Global Generation from a Single Image with Diffusion Models},
  author={Zhang, Xianjie and Zhang, Yusen and He, Yujie and Li, Min},
  journal={Journal of Artificial Intelligence and Soft Computing Research},
  volume={16},
  number={1},
  pages={39--53},
  year={2026},
  publisher={De Gruyter Poland}
}

@inproceedings{wu2024seesr,
  title={Seesr: Towards semantics-aware real-world image super-resolution},
  author={Wu, Rongyuan and Yang, Tao and Sun, Lingchen and Zhang, Zhengqiang and Li, Shuai and Zhang, Lei},
  booktitle={Proceedings of the IEEE/CVF conference on computer vision and pattern recognition},
  pages={25456--25467},
  year={2024}
}

@inproceedings{wang2024sinsr,
  title={Sinsr: diffusion-based image super-resolution in a single step},
  author={Wang, Yufei and Yang, Wenhan and Chen, Xinyuan and Wang, Yaohui and Guo, Lanqing and Chau, Lap-Pui and Liu, Ziwei and Qiao, Yu and Kot, Alex C and Wen, Bihan},
  booktitle={Proceedings of the IEEE/CVF conference on computer vision and pattern recognition},
  pages={25796--25805},
  year={2024}
}

@inproceedings{liang2021swinir,
  title={Swinir: Image restoration using swin transformer},
  author={Liang, Jingyun and Cao, Jiezhang and Sun, Guolei and Zhang, Kai and Van Gool, Luc and Timofte, Radu},
  booktitle={Proceedings of the IEEE/CVF international conference on computer vision},
  pages={1833--1844},
  year={2021}
}

@inproceedings{dong2025tsd,
  title={Tsd-sr: One-step diffusion with target score distillation for real-world image super-resolution},
  author={Dong, Linwei and Fan, Qingnan and Guo, Yihong and Wang, Zhonghao and Zhang, Qi and Chen, Jinwei and Luo, Yawei and Zou, Changqing},
  booktitle={Proceedings of the Computer Vision and Pattern Recognition Conference},
  pages={23174--23184},
  year={2025}
}

@inproceedings{kim2016accurate,
  title={Accurate image super-resolution using very deep convolutional networks},
  author={Kim, Jiwon and Lee, Jung Kwon and Lee, Kyoung Mu},
  booktitle={Proceedings of the IEEE conference on computer vision and pattern recognition},
  pages={1646--1654},
  year={2016}
}

@inproceedings{tai2017image,
  title={Image super-resolution via deep recursive residual network},
  author={Tai, Ying and Yang, Jian and Liu, Xiaoming},
  booktitle={Proceedings of the IEEE conference on computer vision and pattern recognition},
  pages={3147--3155},
  year={2017}
}

@article{depthanything3,
  title={Depth Anything 3: recovering the visual space from any views},
  author={Haotong Lin and Sili Chen and Jun Hao Liew and Donny Y. Chen and Zhenyu Li and Guang Shi and Jiashi Feng and Bingyi Kang},
  journal={arXiv preprint arXiv:2511.10647},
  year={2025}
}

@inproceedings{zhu2024flowie,
  title={Flowie: Efficient image enhancement via rectified flow},
  author={Zhu, Yixuan and Zhao, Wenliang and Li, Ao and Tang, Yansong and Zhou, Jie and Lu, Jiwen},
  booktitle={Proceedings of the IEEE/CVF Conference on Computer Vision and Pattern Recognition},
  pages={13--22},
  year={2024}
}

@inproceedings{liu2022target,
  title={Target-aware dual adversarial learning and a multi-scenario multi-modality benchmark to fuse infrared and visible for object detection},
  author={Liu, Jinyuan and Fan, Xin and Huang, Zhanbo and Wu, Guanyao and Liu, Risheng and Zhong, Wei and Luo, Zhongxuan},
  booktitle={Proceedings of the IEEE/CVF conference on computer vision and pattern recognition},
  pages={5802--5811},
  year={2022}
}

@inproceedings{liu2023multi,
  title={Multi-interactive feature learning and a full-time multi-modality benchmark for image fusion and segmentation},
  author={Liu, Jinyuan and Liu, Zhu and Wu, Guanyao and Ma, Long and Liu, Risheng and Zhong, Wei and Luo, Zhongxuan and Fan, Xin},
  booktitle={Proceedings of the IEEE/CVF international conference on computer vision},
  pages={8115--8124},
  year={2023}
}

@inproceedings{tang2025controlfusion,
  title={ControlFusion: A Controllable Image Fusion Network with Language-Vision Degradation Prompts},
  author={Tang, Linfeng and Wang, Yeda and Cai, Zhanchuan and Jiang, Junjun and Ma, Jiayi},
  booktitle={The Thirty-ninth Annual Conference on Neural Information Processing Systems},
  year={2025}
}

@article{zou2026axispose,
  title={AxisPose: Model-Free Matching-Free Single-Shot 6D Object Pose Estimation via Axis Generation},
  author={Zou, Yang and Qi, Zhaoshuai and Liu, Yating and Liu, Weiyi and Xu, Zihao and Sun, Weipeng and Li, Xingyuan and Yang, Jiaqi and Zhang, Yanning},
  journal={IEEE Transactions on Circuits and Systems for Video Technology},
  year={2026},
  publisher={IEEE}
}

@article{zou2026toward,
  title={Toward Real-world Infrared Image Super-Resolution: A Unified Autoregressive Framework and Benchmark Dataset},
  author={Zou, Yang and Ma, Jun and Jiao, Zhidong and Li, Xingyuan and Jiang, Zhiying and Liu, Jinyuan},
  journal={arXiv preprint arXiv:2603.04745},
  year={2026}
}

@article{zou2026hatir,
  title={HATIR: Heat-Aware Diffusion for Turbulent Infrared Video Super-Resolution},
  author={Zou, Yang and Zhu, Xingyue and Han, Kaiqi and Ma, Jun and Li, Xingyuan and Jiang, Zhiying and Liu, Jinyuan},
  journal={arXiv preprint arXiv:2601.04682},
  year={2026}
}

@inproceedings{zou2024enhancing,
  title={Enhancing neural radiance fields with adaptive multi-exposure fusion: A bilevel optimization approach for novel view synthesis},
  author={Zou, Yang and Li, Xingyuan and Jiang, Zhiying and Liu, Jinyuan},
  booktitle={Proceedings of the AAAI conference on artificial intelligence},
  volume={38},
  number={7},
  pages={7882--7890},
  year={2024}
}

@article{zou2024contourlet,
  title={Contourlet refinement gate framework for thermal spectrum distribution regularized infrared image super-resolution},
  author={Zou, Yang and Chen, Zhixin and Zhang, Zhipeng and Li, Xingyuan and Ma, Long and Liu, Jinyuan and Wang, Peng and Zhang, Yanning},
  journal={International Journal of Computer Vision},
  year={2026}
}

@inproceedings{liu2025toward,
  title={Toward a Training-Free Plug-and-Play Refinement Framework for Infrared and Visible Image Registration and Fusion},
  author={Liu, Yating and Zou, Yang and Li, Xingyuan and Zhu, Xingyue and Han, Kaiqi and Jiang, Zhiying and Ma, Long and Liu, Jinyuan},
  booktitle={Proceedings of the 33rd ACM International Conference on Multimedia},
  pages={1268--1277},
  year={2025}
}

@article{du2026unsupervised,
  title={Unsupervised hyperspectral image super-resolution via self-supervised modality decoupling},
  author={Du, Songcheng and Zou, Yang and Wang, Zixu and Li, Xingyuan and Li, Ying and Shang, Changjing and Shen, Qiang},
  journal={International Journal of Computer Vision},
  volume={134},
  number={4},
  pages={152},
  year={2026},
  publisher={Springer}
}

\end{document}